\documentclass[10pt,twocolumn,letterpaper]{article}

\usepackage{cvpr}
\usepackage{times}
\usepackage{epsfig}
\usepackage{graphicx}
\usepackage{amsmath}
\usepackage{amssymb}
\usepackage{caption}
\usepackage{url}
\usepackage[breaklinks=true,bookmarks=false]{hyperref}

\cvprfinalcopy 

\def\cvprPaperID{108} 

\ifcvprfinal\fi

\begin{document}

\title{WiCV@CVPR2024: The Thirteenth Women In Computer Vision Workshop at the Annual CVPR Conference}

\author{Asra Aslam$^1$, Sachini Herath$^2$, Ziqi Huang$^3$, Estefania Talavera$^4$,\\ Deblina Bhattacharjee$^5$, Himangi Mittal$^6$, Vanessa Staderini
$^7$, Mengwei Ren$^8$, Azade Farshad$^9$ \\\\ $^1$University of Leeds, $^2$Simon Fraser University, $^3$Nanyang Technological University,\\$^4$University of Twente, $^5$University of Bath, $^6$Carnegie Mellon University, \\ $^7$Austrian Institute of Technology, $^8$Adobe, $^9$Technical University of Münich\\
 \tt\small wicvcvpr2024-organizers@googlegroups.com
}
\maketitle

\begin{abstract}
\thispagestyle{empty}
In this paper, we present the details of Women in Computer Vision Workshop - WiCV 2024, organized alongside the CVPR 2024 in Seattle, Washington, United States.  WiCV aims to amplify the voices of underrepresented women in the computer vision community, fostering increased visibility in both academia and industry. We believe that such events play a vital role in addressing gender imbalances within the field. The annual WiCV@CVPR workshop offers a)~opportunity for collaboration between researchers from minority groups, b)~mentorship for female junior researchers,  c)~financial support to presenters to alleviate financial burdens and d)~a diverse array of role models who can inspire younger researchers at the outset of their careers. In this paper, we present a comprehensive report on the workshop program, historical trends from the past WiCV@CVPR events, and a summary of statistics related to presenters, attendees, and sponsorship for the WiCV 2024 workshop.

\end{abstract}

\section{Introduction}
Despite remarkable progress in various computer vision research areas in recent years, the field still grapples with a persistent lack of diversity and inclusion. While the field of computer vision rapidly expands, female researchers remain underrepresented in the area, constituting only a small amount of professionals in both academia and industry. Due to this, many female computer vision researchers can feel isolated in workspaces which remain unbalanced due to the lack of inclusion.

The WiCV workshop is a gathering designed for all individuals, irrespective of gender, engaged in computer vision research. It aims to appeal to researchers at all levels, including established researchers in both industry and academia (e.g. faculty or postdocs), graduate students pursuing a Masters or PhD, as well as undergraduates interested in research.  The overarching goal is to enhance the visibility and recognition of female computer vision researchers across these diverse career stages, reaching women from various backgrounds in educational and industrial settings worldwide.

There are three key objectives of the WiCV workshop:

\paragraph{Networking and Mentoring}

The first objective is to expand the WiCV network and facilitate interactions between members of this network. This includes female students learning from seasoned professionals who share career advice and experiences. A mentoring banquet held alongside the workshop provides a casual environment for junior and senior women in computer vision to meet, exchange ideas and form mentoring or research relationships.

\paragraph{Raising Visibility}

The workshop's second objective is to elevate the visibility of women in computer vision, both at junior and senior levels. Senior researchers are invited to give high quality keynote talks on their research, while junior researchers are encouraged to submit their recent or ongoing work, with many of these being selected for oral or poster presentation through a rigorous peer review process. This empowers junior female researchers to gain experience presenting their work in a professional yet supportive setting. The workshop aims for diversity not only in research topics but also in the backgrounds of presenters. Additionally, a panel discussion provides a platform for female colleagues to address topics of inclusion and diversity.

\paragraph{Supporting Junior Researchers}

Finally, the third objective is to offer junior female researchers the opportunity to attend a major computer vision conference that might otherwise be financially inaccessible. This is made possible through travel grants awarded to junior researchers who present their work during the workshop's poster session. These grants not only enable participation in the WiCV workshop but also provide access to the broader CVPR conference.

\section{Workshop Program}
\label{program}
The workshop program featured a diverse array of sessions, including 4 keynotes, 6 oral presentations, 41 poster presentations, a panel discussion, and a mentoring session. Consistent with previous years, our keynote speakers were carefully selected to ensure diversity in terms of the topics that were covered, their backgrounds, whether they work in academia or industry, and their seniority.  This deliberate choice of diverse speakers is of paramount importance, as it offers junior researchers a multitude of potential role models with whom they can resonate and, in turn, envision their unique career paths.

The workshop schedule at CVPR 2024 featured a diverse range of sessions and activities, including:
\begin{itemize}
    \item Introduction
    
    \item Invited Talk 1: Guoying Zhao (Academy Finland and the University of Oulu), \textit{Computer Vision in Affective Computing}
    
    \item Oral Session 1
    \begin{itemize}
        \item Gemma Canet Tarrés, \textit{PARASOL: Parametric Style Control for Diffusion Image Synthesis}
        \item Xinye Wanyan, \textit{Extending global-local view alignment for self-supervised learning with remote sensing imagery}
    \end{itemize}
    
    \item Invited Talk 2: Shek Azizi (Google DeepMind), \textit{Generalist Biomedical AI: Towards Scalable Healthcare Impact}
    
    \item Sponsors Exhibition (in person) for Apple and Wayve.
    
    \item Oral Session 2
    \begin{itemize}
        \item Mehwish Mehmood, \textit{RetinaLiteNet: A Lightweight Transformer based CNN for Retinal Feature Segmentation}
        \item Taiba Majid, \textit{ABC-CapsNet: Attention based Cascaded Capsule Network for Audio Deepfake Detection}
    \end{itemize}
    
    \item Invited Talk 3: Elisa Ricci (University of Trento), \textit{Harnessing Language for Video Understanding without Training}
    
    \item Sponsors Exhibition (in person) for Toyota Research and Meta.
    
    \item Oral Session 3
    \begin{itemize}
        \item Mallika Garg, \textit{GestFormer: Multiscale Wavelet Pooling Transformer Network for Dynamic Hand Gesture Recognition}
        \item Yingchao Huang, \textit{Unsupervised Domain Adaptation for Weed Segmentation Using Greedy Pseudo-labelling}
    \end{itemize}
    
    \item Invited Talk 4: Boyi Li (UC Berkeley \& NVIDIA Research), \textit{Vision and Language for Interactive Robot Task Planning}
    
    \item Panel Discussion by Abby Stylianou, Angel Chang, Devi Parikh, Ilke Demir, Judy Hoffman, and Kristen Grauman
    
    \item Poster Session (in person as well as virtual)
    
    \item Closing Remarks
    
    \item Mentoring Session, Talks, and Dinner (in person)
    \begin{itemize}       
        \item Invited talk:
        \begin{enumerate}
            \item Kate Saenko (FAIR Labs, Meta and Full Professor, Boston University)
            \item Cornelia Fermüller (Autonomy Cognition and Robotics Lab, UMD) on \textit{Music Education for All}
        \end{enumerate}

         \item In-Person mentoring session and dinner: Boyi Li, Ana-Maria Marcu, Elisa Ricci, Khadija Khaldi, Djamila Aouada, Mingfei Yan, Petia Radeva, Syenny, Gul Varol, Mahdieh Poostchi, Shek Azizi, Melissa Hall, Guoying Zhao, Kiana ehsani, Kate Saneko, Cornellia Fermüller, Katherine Liu, Ilke Demir, Caroline Pantofaru, Sofía Josefina Lago Dudas, Sasha Harrison, Diane Zhu, Yujiao Shi, and Tanya Glozman.
         
        \item Virtual mentoring session on Zoom.
    \end{itemize}
\end{itemize}

\subsection{Hybrid Setting}
This year, our organizational approach underwent slight adjustments due to CVPR 2024 being held in a hybrid setting, accommodating both in-person and virtual attendance. We took deliberate steps to sure that the virtual WiCV workshop was an engaging and interactive event. To achieve this, we took the following steps: Talks, oral sessions, and the panel were streamed via Zoom for virtual attendances. The poster session was repeated virtually a week after the conference, mirroring the format of the main conference. We also facilitated online mentoring sessions via Zoom, catering to mentors and mentees who could only participate virtually. 

\section{Workshop Statistics}

The first edition of the Women in Computer Vision (WiCV) workshop was held in conjunction with CVPR 2015. Over the years, both the participation rate and the quality of submissions to WiCV have steadily increased.  
Following the examples from the editions held in previous years \cite{asra2023wicv, antensteiner2022wicv, doughty2021wicv, Amerini19, Akata18, Demir18} we have continued to curate top-quality submissions into our workshop proceedings. By providing oral and poster presenters the opportunity to publish their work in the conference's proceedings, we aim to further boost the visibility of female researchers. 

This year, the workshop was held as a half-day in-person event with hybrid options, while the virtual component was hosted via Topia and Zoom. The in-person gathering took place at the Seattle Convention Center in Seattle, Washington, United States. Senior and junior researchers were invited to present their work, including the poster presentations detailed in the previous Section \ref{program}.\\

The organizers for this year's WiCV workshop come from diverse backgrounds in both academia and industry, representing various institutions across different time zones. Their diverse backgrounds and wide-ranging research areas have enriched the organizing committee's perspectives and contributed to a well-rounded approach. Their broad range of research interests in computer vision and machine learning encompass video understanding, object detection, non-verbal communication, open-source benchmark datasets, activity recognition, anomaly detection, autoencoders, generalization, captioning, 3D Point Cloud, medical imaging and vision for robotics.

\begin{figure}[h]
\centering
\includegraphics[width=1\linewidth]{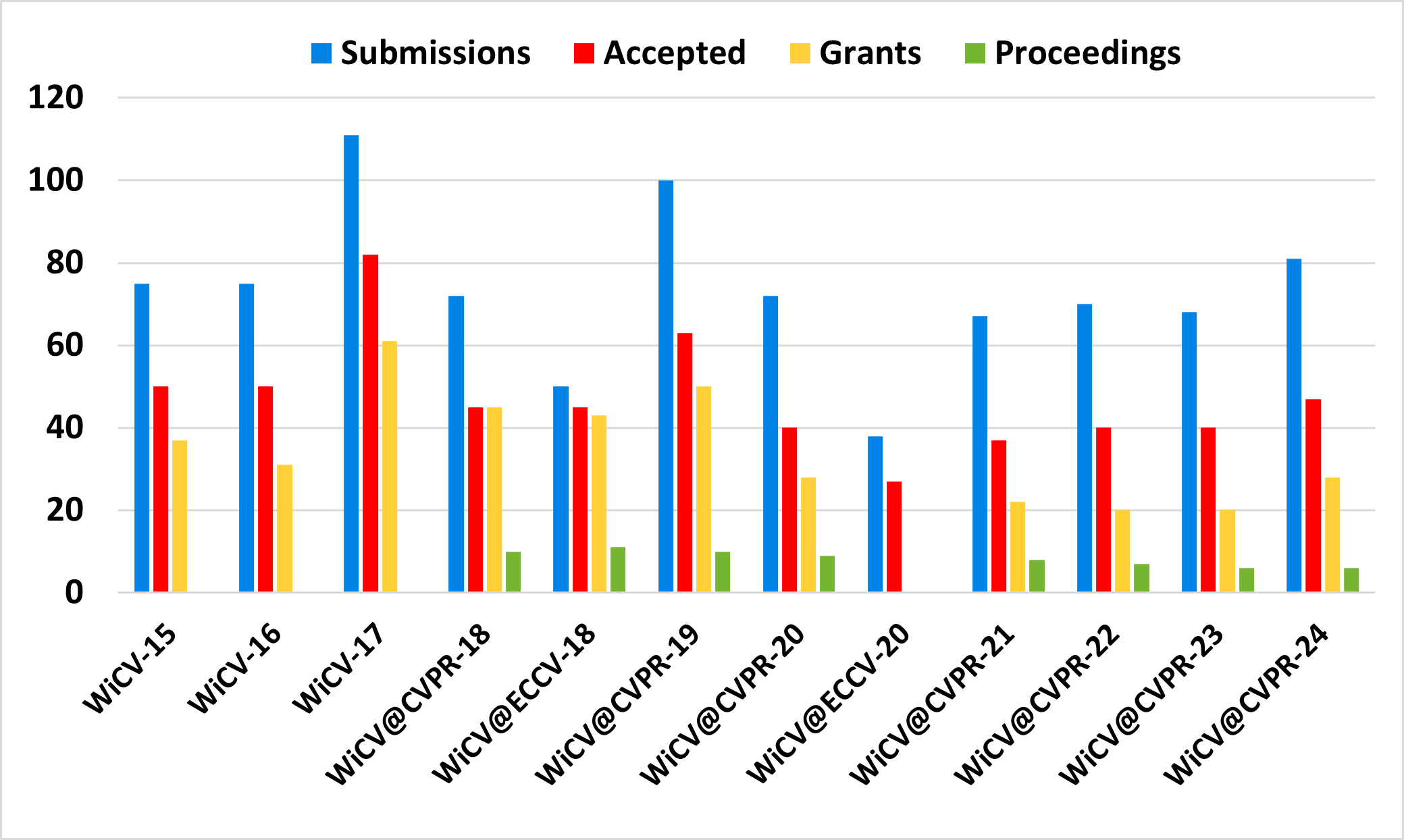}
\captionof{figure}{\textbf{WiCV Submissions.} The number of submissions over the past years of WiCV.}
\label{fig:sub}
\end{figure}

This year, we received 81 high-quality submissions from a wide range of topics and institutions, which is on par with WiCV@CVPR23. The most popular topics included deep learning architectures and techniques followed by 2D and 3D Computer Vision, Transformers, Diffusion, Segmentation, Classification, and Medical applications. Out of the 81 submissions, 65 underwent the review process. Six papers were selected as oral presentations and inclusion in the CVPR24 workshop proceedings, while 41 papers were chosen for poster presentations. The comparison with previous years is presented in Figure~\ref{fig:sub}. Thanks to the diligent efforts of an interdisciplinary program committee comprising 41 reviewers, the submitted papers received thorough evaluations and valuable feedback. Additionally, during the mentoring session, 48 mentees received in-person guidance from 8 mentors, and 5 mentees attended virtual sessions with 2 mentors in separate meetings via Zoom.

This year, we continued the WiCV tradition from previous workshops \cite{Akata18,Amerini19,Demir18,doughty2021wicv, goel2022wicv} by providing grants to assist the authors of accepted submissions in participating in the workshop. These grants covered a range of expenses, with the specifics varying for each attendee depending on their individual needs, including, for example, conference registration fees, round-trip flight itineraries, and two days of accommodation for all authors of accepted submissions who requested funding.

\begin{figure}
\centering
\includegraphics[width=1\linewidth]{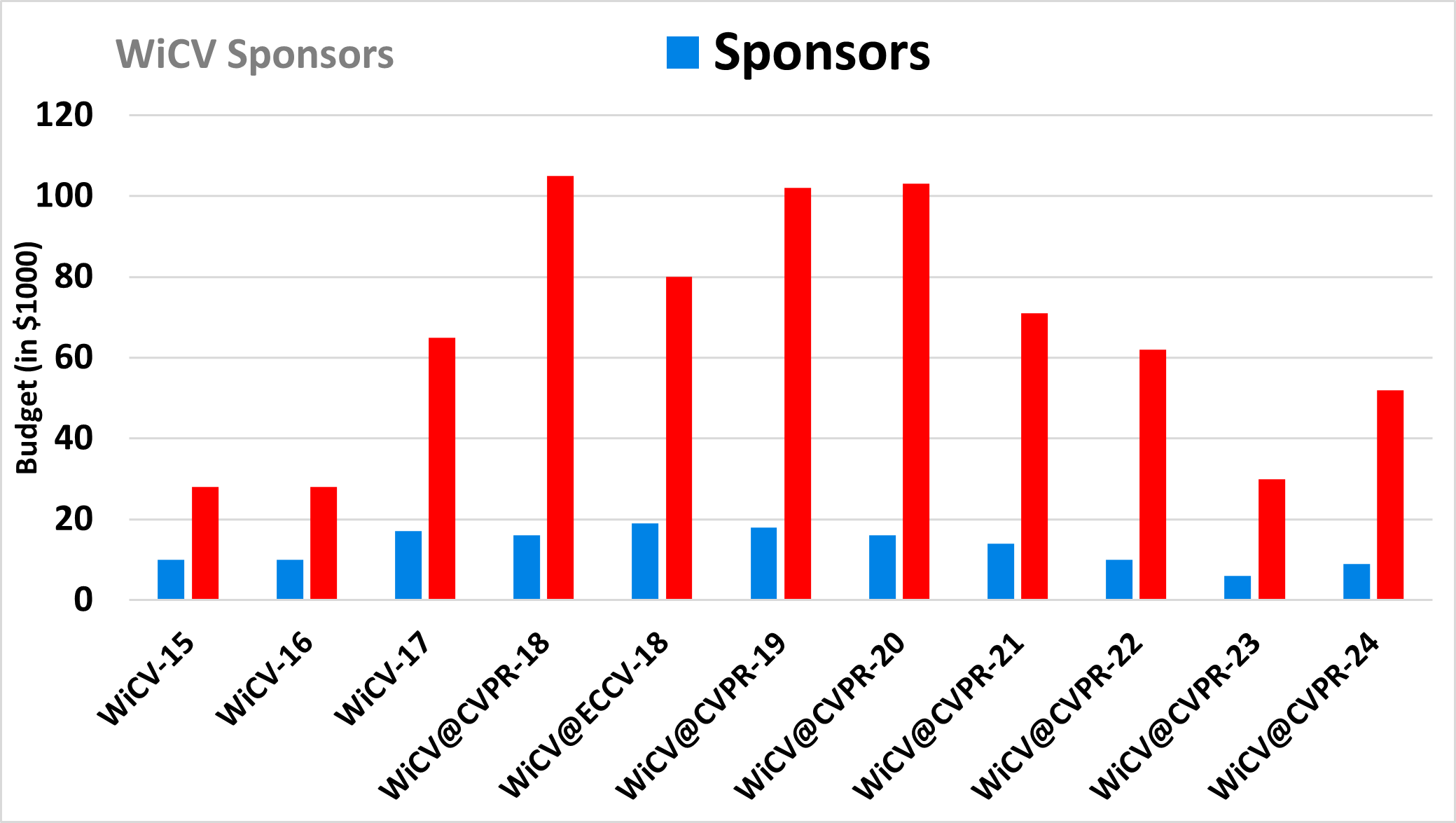}
\captionof{figure}{\textbf{WiCV Sponsors.} The number of sponsors and the amount of sponsorship for WiCV. The amount is expressed in US dollar (USD).}
\label{fig:spo}
\end{figure}

The total sponsorship for this year's workshop amounted to \$52,000 USD, with contributions of 9 sponsors, meeting our target. In Figure~\ref{fig:spo} you can find the details with respect to the past years. 

\section{Conclusions}
WiCV at CVPR 2024 has once again proven to be a valuable opportunity for presenters, participants, and organizers, providing a platform to unite the community. It continues to address the persistent issue of gender balance in our field, and we believe it has played a significant role in strengthening the community. It provided an opportunity for people to connect from all over the world from the comfort of their personal spaces. With a high number of paper submissions and even greater number of attendees, we anticipate that the workshop will continue the positive trajectory of previous years, fostering a stronger sense of community, increased visibility, and inclusive support and encouragement for all female researchers in academia and industry. Moreover, WiCV Members participated in the Diversity \& Inclusion Social event at CVPR. Furthermore, WiCV also got featured by in the CVPR~2024 Magazine \cite{cvprmagazine} and Diversity, Equity, \& Inclusion  News \cite{cvprnews}.

\section{Acknowledgments}
We express our sincere gratitude to our sponsors, including our Platinum sponsors: Apple, Wayve, Meta, and Toyota Research Institute, as well as our Silver Sponsor: google Research, and Bronze sponsors: Meshcapade, Austrian Institute of Technology, Disney Research, and Tencent. Our appreciation also extends to the San Francisco Study Center, our fiscal sponsor, for their invaluable assistance in managing sponsorships and travel awards. We are thankful for the support and knowledge-sharing from organizers of previous WiCV workshops, without whom this WiCV event would not have been possible. Finally, we extend our heartfelt thanks to the dedicated program committee, authors, reviewers, submitters, and all participants for their valuable contributions to the WiCV network community.

\section{Contact}
\noindent \textbf{Website}: \url{https://sites.google.com/view/wicv-cvpr-2024/}\\
\textbf{E-mail}: wicvcvpr2024-organizers@googlegroups.com\\
\textbf{Facebook}: \url{https://www.facebook.com/WomenInComputerVision/}\\
\textbf{Twitter}: \url{https://twitter.com/wicvworkshop}\\
\textbf{Google group}: women-in-computer-vision@googlegroups.com\\

{\small
\bibliographystyle{ieee}
\bibliography{egbib}
}

\end{document}